\documentclass{article}
\usepackage{graphicx}
\usepackage{amsmath}
\usepackage{amssymb}
\usepackage{algorithm}
\usepackage{algorithmic}
\usepackage{url}
\usepackage{color}
\usepackage[breaklinks=true,colorlinks]{hyperref}

\definecolor{Myblue}{rgb}{0,.3,.6}
\newcommand{\emc}[1]{{\textbf{\textit{\color{Myblue}#1}}}}

\DeclareMathOperator{\diag}{diag}

\DeclareMathOperator{\erf}{erf}

\DeclareMathOperator{\sign}{sign}

\DeclareMathOperator{\vecc}{vec}

\newcommand{\qed}{\hfill \ensuremath{\Box}} 
\newenvironment{proof}[1][Proof]{\begin{trivlist}
\item[\hskip \labelsep {\bfseries #1}]}{\end{trivlist}}

\newcommand{\linesep}{\begin{center} {\color{red}\line(1,0){300}} \end{center} \vspace{-0.4in} \begin{center} {\color{red}\line(1,0){250}} \end{center}}

\title{Training Recurrent Neural Networks by Diffusion}
\date{}

\author{Hossein Mobahi \\
Computer Science \& Artificial Intelligence Lab.\\
Massachusetts Institute of Technology\\
Cambridge, MA, USA \\
\color{magenta}\texttt{hmobahi@csail.mit.edu} \\
}

\begin{document}

\maketitle

\begin{abstract}
This work presents a new algorithm for training recurrent neural networks (although ideas are applicable to feedforward networks as well). The algorithm is derived from a theory in nonconvex optimization related to the diffusion equation. The contributions made in this work are two fold. First, we show how some seemingly disconnected mechanisms used in deep learning such as smart initialization, annealed learning rate, layerwise pretraining, and noise injection (as done in dropout and SGD) arise naturally and automatically from this framework, without manually crafting them into the algorithms. Second, we present some preliminary results on comparing the proposed method against SGD. It turns out that the new algorithm can achieve similar level of generalization accuracy of SGD in much fewer number of epochs.
\end{abstract}

\section{Introduction}

Deep learning has recently beaten records in image recognition \cite{AlexNet}, speech recognition \cite{HintonSpeech} and has made significant improvements in natural language processing \cite{BahdanauCB14,Sut}. However, currently ``training'' deep networks, and specially recurrent neural networks (RNNs), is a challenging task \cite{RNNHard}. To improve learning (in terms of convergence speed, attained training cost and generalization error) gradient based optimization methods are often used in combination with other techniques such as smart initialization \cite{Init13}, layerwise pretraining \cite{Bengio06}, dropout \cite{dropout}, annealed learning rate, and curriculum learning \cite{Beng09Cur}.

\begin{figure}
\centering \includegraphics[width=.4\textwidth,height=.3\textwidth]{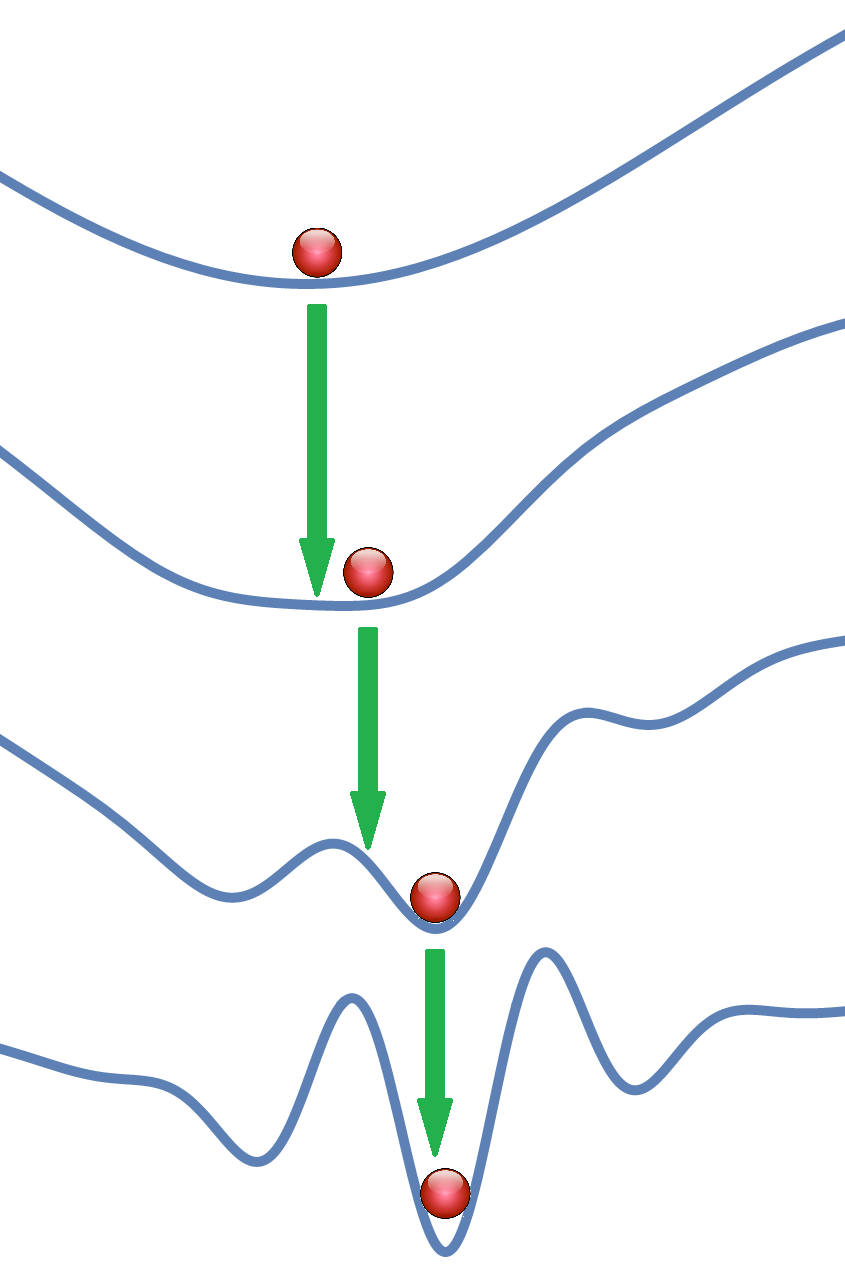}
\caption{Optimization by the continuation method. Top is the simplified function and bottom is the original complex objective function. The solution of each subproblem initializes the subproblem below it.}
\label{fig:continuation}
\end{figure}

The difficulty in training deep networks is mainly attributed to their optimization landscape, where saddle points \cite{Dauphin,PascanuDGB14}, plateaus, and sharp curvatures are prevalent. A general strategy for tackling difficult optimization problems is the \emc{continuation method}. This method gradually transforms a highly simplified version of the problem back to its original form while following the solution along the way. The simplified problem is supposedly easy to solve. Then, each intermediate subproblem is initialized by the solution from the previous subproblem until reaching the final problem (see Figure \ref{fig:continuation}).

There are two loose ends for using optimization by continuation: 1. how to choose the simplified problem, 2. how to transform the simplified problem to the main task. For both of these questions, there are infinite answers. More precisely, given an objective function, there are infinite ways infinite smooth convex functions that could be used as initial ``easy'' task, and also infinite ways to gradually transform that to the main objective function. The quality of the solution attained by the continuation method \emc{critically depends} on these choices. Recently we have proved that these choices can be made optimally via the \emc{diffusion equation} \cite{MobahiEMMCVPR}. Specifically, the objective function is considered as the initial heat distribution on a domain, and the heat is diffused over time according to the \emc{heat equation}.

The solution to the heat equation on $\mathbb{R}^n$ is known analytically: it is the \emc{convolution} of the initial heat distribution (i.e., the objective function) with the \emc{Gaussian kernel}. Obviously, convolution with the Gaussian kernel smoothes the objective function\footnote{This happens when the objective function has well-defined Fourier transform. Then the convolution transform to product in the frequency domain. As the Fourier transform of the Gaussian is also a Gaussian, the resulted product attenuates higher frequencies.}. The bandwidth parameter $\sigma$ of the Gaussian kernel determines the amount of smoothing. The algorithm for optimization by diffusion starts from a large $\sigma$ (highly simplified objective function), and then follows the minimizer as $\sigma$ shrinks toward zero (which leads to the original cost function).

The optimality result we derived in \cite{MobahiEMMCVPR} is a stepping stone for developing practical algorithms. Specifically, it suggests using Gaussian convolution for creating intermediate optimization tasks, but it does not answer whether the resulted convolution could be computed efficiently or not. In fact, the answer to this question is problem specific. We have shown that for some family of functions such as multivariate polynomials, the resulted convolution can be computed in \emc{closed form} \cite{mobahiclosed}. In this work, we push that result further and show that, up to very reasonable approximation, common objective functions arising in deep learning also have a closed form Gaussian convolution. This is surprising because such objective function is highly nonlinear; involving a nested form of ill-behaved activation functions as such $\sign$ and ReLU.

By studying deep learning through the diffusion and continuation method, we discover two interesting observations. First, from theoretical viewpoint, some common and successful techniques to improve learning, such as noise injection \cite{dropout}, layerwise pretraining \cite{Bengio06}, and annealed learning rate, \emc{automatically emerge} from the diffused cost function. Therefore, this theory \emc{unifies} some seemingly isolated techniques. Second, from a practical viewpoint, training deep networks by this method seems to result in a significant speed up compared to stochastic gradient descent (SGD) method. The preliminary results presented in this draft indicate up to \emc{$25\%$ reduction in training time} for learning RNNs.

This article is organized as follows. We first show that the diffused form of common activation functions has a closed form expression. After that, when we compute the diffused cost function for training a deep network, where the result depends on the diffused activation function introduced earlier. We discuss some properties of the diffused cost function and make connections to noise injection \cite{dropout}, layerwise pretraining \cite{Bengio06}, and annealed learning rate. We conclude this article by presenting a preliminary evaluation of the proposed algorithm against SGD.

\section{Optimization by Diffusion and Continuation}

The optimality of using the diffusion equation for creating intermediate optimization problems is studied in our earlier work \cite{MobahiEMMCVPR}. Briefly, diffusion is a relaxation of a time evolution process that converts an objective function to its convex envelope\footnote{The convex envelope of a function is an interesting choice (versus any other convex function) for the initial simplified version of it for various reasons. 1. Any global minimizer of the cost function is also a global minimizer of its convex envelope. 2. it provides the tightest convex underestimator of the cost function. 3. Geometrically, tt is the function whose epigraph coincides with the convex hull of the epigraph of the cost function.} \cite{Vese}. The latter is a \emc{nonlinear} partial differential equation that lacks a closed form, but once \emc{linearized}, the heat equation (a special type of diffusion equation) arises,

\begin{equation}
\label{eq:diffusion}
\frac{d}{dt} g(\boldsymbol{x},t) = \Delta_{\boldsymbol{x}} g(\boldsymbol{x},t) \quad,\quad \mbox{s.t. } g(\boldsymbol{x},0)=f(\boldsymbol{x}) \,.
\end{equation}

\begin{figure}
\centering
\includegraphics[width=.3\textwidth]{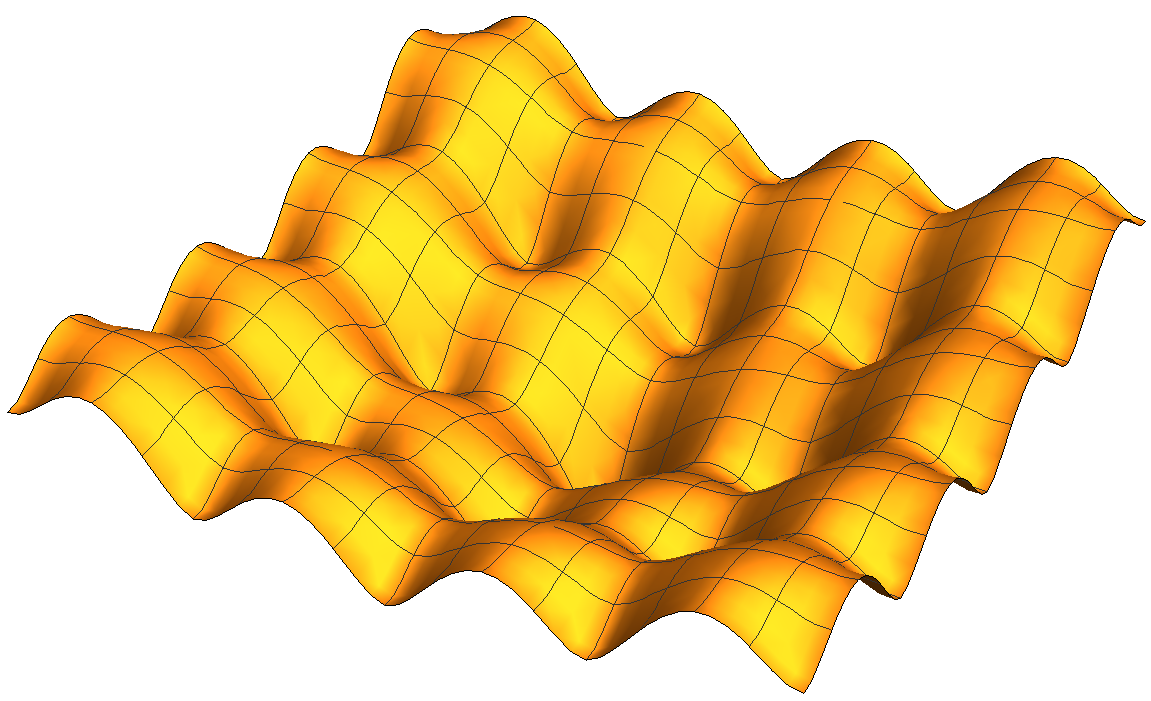}
\includegraphics[width=.3\textwidth]{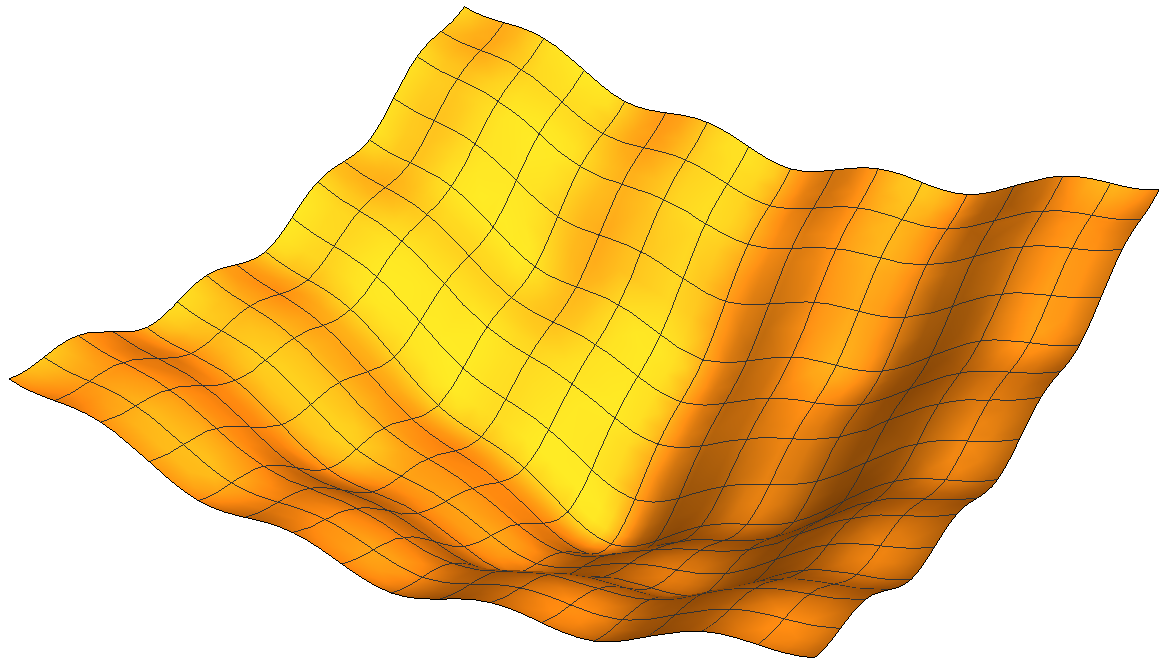}
\includegraphics[width=.3\textwidth]{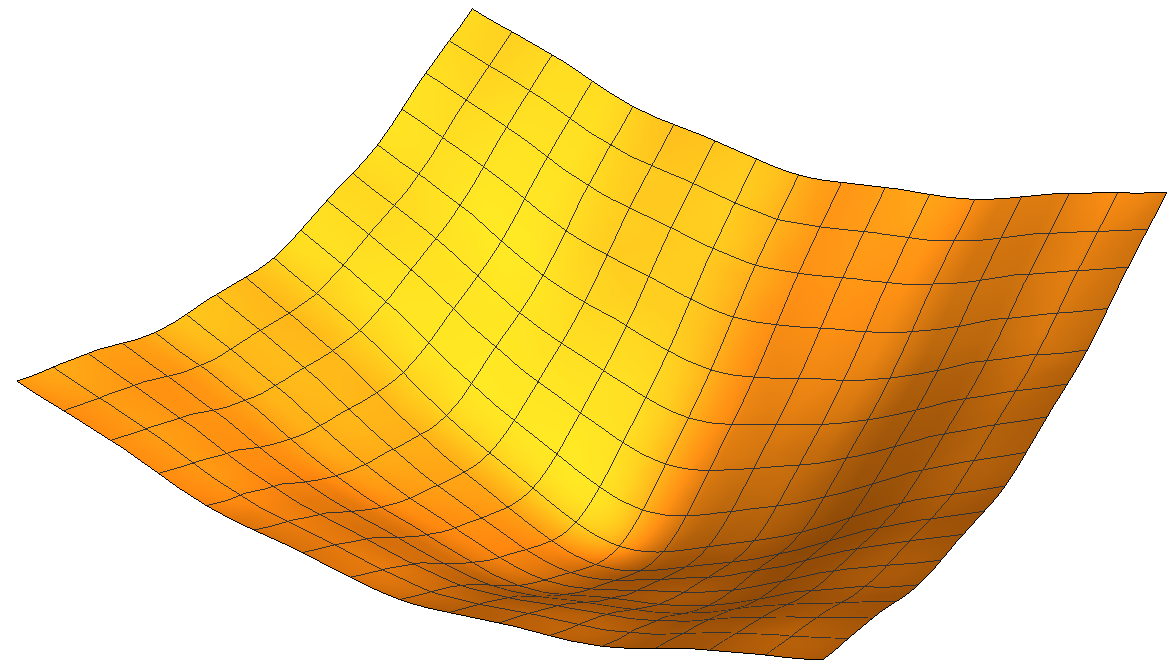}
\caption{Diffusion of Ackley's function with time progressing from the left to the right plot.}
\label{fig:ackley}
\end{figure}

Here $f$ is the original objective function, and $g$ is its time evolution according to the heat equation. Here $\Delta_{\boldsymbol{x}}$ is the Laplace operator w.r.t. the variable $\boldsymbol{x}$. Diffusion is a powerful tool for simplifying the objective function. For example, the number of local minima in the Ackley's function \cite{ackley} is exponential in the number of variables. By diffusing this function via the heat equation, however, all local minima eventually disappear (see Figure \ref{fig:ackley}).

Going from the nonlinear PDE of \cite{Vese} to the (linear) heat equation is computationally of great value. That is, the solution to the heat equation is known analytically \cite{Widder75}: it is the Gaussian convolution of the original (objective) function and the bandwidth parameter of the Gaussian determines the time point at which the diffused function is evaluated. Diffusion combined with the path following lead to a simple optimization algorithm listed in Algorithm \ref{alg:alg_goal}.

\begin{algorithm} [t]
\caption{Algorithm for Optimization by Diffusion and Continuation}
\label{alg:alg_goal}
\begin{algorithmic} [1]
\STATE Input: $f:\mathcal{X} \rightarrow \mathbb{R}$, Sequence $\infty>\sigma_0>\sigma_1>\dots>\sigma_m = 0$.
\STATE $\boldsymbol{x}_0=$ global minimizer of $g(\boldsymbol{x};\sigma_0)$.
\FOR {{$k=1$} \textbf{to} {$m$}}
\STATE $\boldsymbol{x}_{k}=$ Local minimizer of $g(\boldsymbol{x};\sigma_k)$, initialized at $\boldsymbol{x}_{k-1}$.
\ENDFOR
\STATE Output: $\boldsymbol{x}_m$
\end{algorithmic}
\end{algorithm}

\section{Diffused Activation Functions}

Let $k_\sigma(\boldsymbol{w})$ be the Gaussian kernel with zero mean and covariance $\sigma^2 \boldsymbol{I}$. The diffused activation functions listed in Table \ref{tab:Table} are simply obtained\footnote{All listed diffused functions are exact except $\tanh$. Unfortunately, $\tanh \star k_\sigma$ does not have a closed form. We leverage the approximation $\tanh(y) \approx \erf(\frac{\sqrt{\pi}}{2} y)$. Notice that we know the exact diffused form for $\erf$ as listed in the table. Thus, by convolving both sides with $k_\sigma$ we obtain $[\tanh \star k_\sigma] (y) \approx \erf(\frac{\sqrt{\pi}}{2} \frac{y}{\sqrt{1+\frac{\pi}{2} \sigma^2}})$. The R.H.S. of the latter form can be again approximated via $\tanh(y) \approx \erf(\frac{\sqrt{\pi}}{2} y)$. This leads to the approximate identity $[\tanh \star k_\sigma] (y) \approx \tanh( \frac{y}{\sqrt{1+\frac{\pi}{2} \sigma^2}})$.} by convolving them with the Gaussian $k_\sigma$. Similar forms of smoothed ReLU and $\sign$ are used by \cite{jordan} with a fixed $\sigma=\frac{1}{\sqrt{2 \pi}}$, for a proving learnability of deep networks.

\begin{table}
\begin{center}
\begin{tabular}{|l|l|l|}
\hline
\bf Name & \bf Original & \bf Diffused \\
\hline
Sign & $\sign(x)$ & $\erf(\frac{x}{\sqrt{2}\sigma})$ \\
\hline
Error & $\erf(a x)$ & $\erf(\frac{a x}{\sqrt{1+2 (a \sigma)^2}})$ \\
\hline
Tanh & $\tanh(x)$ & $\tanh( \frac{x}{\sqrt{1 +\frac{\pi}{2}\sigma^2 } })$ \\
\hline
ReLU & $\max(0,x)$ & $\frac{\sigma}{\sqrt{2 \pi}} e^{-\frac{x^2}{2 \sigma^2} } + \frac{1}{2} x \big (1+\erf(\frac{x}{\sqrt{2}\sigma}) \big)$ \\
\hline
\end{tabular}
\caption{List of some functions and their diffused form by the heat kernel.}
\end{center}
\label{tab:Table}
\end{table}

\begin{center}
\begin{tabular}{c c c}
\includegraphics[width=.32\textwidth]{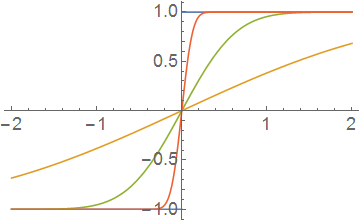} &
\includegraphics[width=.32\textwidth]{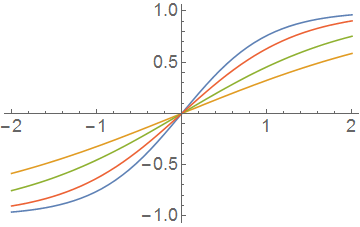} &
\includegraphics[width=.32\textwidth]{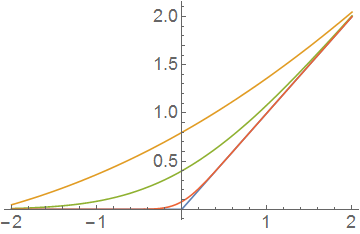} \\
$Sign$ & $Tanh$ & $ReLU$
\end{tabular}
\end{center}
{\footnotesize{Plot of smoothed responses of activation functions within $x \in [-2,2]$. Blue is the original function. Red, green, and orange show the suggested functions with $\sigma_{\mbox{red}}< \sigma_{\mbox{grn}} <\sigma_{\mbox{orn}} $.}}

\section{Training RNNs}

\subsection{RNN Cost Function}

Given a set of $S$ training sequences, each of length $T$. Denote the $s$'th sequence by $\langle (\boldsymbol{x}_{s,1},\boldsymbol{y}_{s,1}), \dots, (\boldsymbol{x}_{s,T},\boldsymbol{y}_{s,T}) \rangle$. Given some discrepancy function $d$. The problem of sequence learning by an RNN can be stated as below,

\begin{eqnarray}
& & \min_{\boldsymbol{a},\boldsymbol{b}, \boldsymbol{m}_0, \boldsymbol{U},\boldsymbol{V}, \boldsymbol{W}} \sum_{s=1}^S \sum_{t=1}^T d(h(\boldsymbol{n}_{s,t})-\boldsymbol{y}_{s,t}) \\
\mbox{s.t.} & & \boldsymbol{n}_{s,t} \triangleq \boldsymbol{W} \, h(\boldsymbol{m}_{s,t} ) + \boldsymbol{b} \\
& & \boldsymbol{m}_{s,t} \triangleq \boldsymbol{U} \boldsymbol{x}_{s,t} + \boldsymbol{V} h(\boldsymbol{m}_{s,t-1}) \, + \boldsymbol{a} \,,
\end{eqnarray}

where $\boldsymbol{a}$, $\boldsymbol{b}$, $\boldsymbol{m}_0$, $\boldsymbol{W}$, $\boldsymbol{U}$ and $\boldsymbol{V}$ are the weights of the network. Denote the dimension of $\boldsymbol{x}_{s,t}$ and $\boldsymbol{y}_{s,t}$ be $X$ and $Y$ respectively. Also denote the number of neurons by $H$. Then, $\boldsymbol{a}$ is $H \times 1$, $\boldsymbol{b}$ is $Y \times 1$, $\boldsymbol{m}_0$ is $H \times 1$, $\boldsymbol{W}$ is $Y \times H$, $\boldsymbol{U}$ is $H \times X$, and $\boldsymbol{V}$ is $H \times H$. Obviously $\boldsymbol{n}_{s,t}$ is $Y \times 1$ and $\boldsymbol{m}_{s,t}$ is $H \times 1$.

Suppose $\boldsymbol{m}_{s,0}=\boldsymbol{m}_0$, i.e. the initial state is independent of the training sequence. Here $h$ is some activation function. When the argument of $h$ is a vector, the result will be a vector of the same size, whose entries consists of the element-wise application of $h$.

\begin{figure}
\begin{center}
\includegraphics[width=1\textwidth]{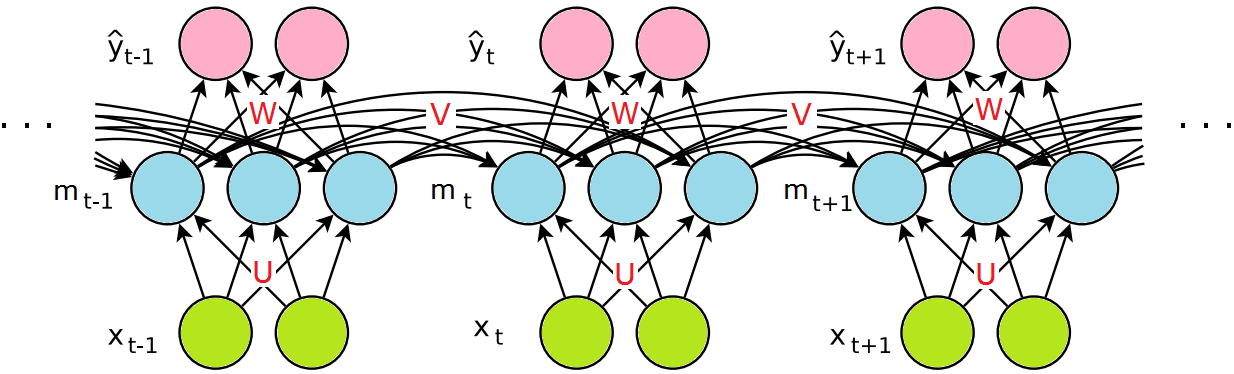}
\end{center}
\caption{A Recurrent Neural Network. Figure is adapted with permission from \cite{RNNHard} and slightly modified.}
\end{figure}

Treating each $\boldsymbol{n}_{s,t}$ and $\boldsymbol{m}_{s,t}$ as \emc{independent variables} and forcing their definition (equality) by some penalty function, we arrive at the following \emc{unconstrained} problem,

\begin{eqnarray}
\min_{\boldsymbol{a},\boldsymbol{b}, \boldsymbol{m}_0, \boldsymbol{U},\boldsymbol{V}, \boldsymbol{W}, \boldsymbol{M}, \boldsymbol{N}} \sum_{s=1}^S \sum_{t=1}^T & & d(h(\boldsymbol{n}_{s,t})-\boldsymbol{y}_{s,t}) \nonumber \\
&+& \lambda \Big( p \big( \boldsymbol{W} \, h(\boldsymbol{m}_{s,t} ) + \boldsymbol{b} - \boldsymbol{n}_{s,t} \big) \,+\,p \big( \boldsymbol{U} \boldsymbol{x}_t + \boldsymbol{V} h(\boldsymbol{m}_{s,t-1}) \, + \boldsymbol{a} - \boldsymbol{m}_{s,t} \big) \Big)\,, \nonumber
\end{eqnarray}

where the notation $\boldsymbol{N}$ and $\boldsymbol{M}$ are matrices whose columns are comprised of $\boldsymbol{n}_{s,t}$ and $\boldsymbol{m}_{s,t}$ for all choices of $(s,t)$.

Letting, $d(\boldsymbol{e}) \triangleq \| \boldsymbol{e}\|^2$ (mean squared error) and $p(\boldsymbol{e}) \triangleq \| \boldsymbol{e}\|^2$ (quadratic penalty), the problem can be expressed as below,

\begin{eqnarray}
\min_{\boldsymbol{a},\boldsymbol{b}, \boldsymbol{m}_0, \boldsymbol{U},\boldsymbol{V}, \boldsymbol{W}, \boldsymbol{M}, \boldsymbol{N}} \sum_{s=1}^S \sum_{t=1}^T & & \| h(\boldsymbol{n}_{s,t})-\boldsymbol{y}_{s,t} \|^2 \nonumber \\
&+& \lambda \big( \| \boldsymbol{W} \, h(\boldsymbol{m}_{s,t} ) + \boldsymbol{b} - \boldsymbol{n}_{s,t} \|^2 \,+\, \| \boldsymbol{U} \boldsymbol{x}_{s,t} + \boldsymbol{V} h(\boldsymbol{m}_{s,t-1}) \, + \boldsymbol{a} - \boldsymbol{m}_{s,t} \|^2 \big)\,. \nonumber
\end{eqnarray}

Here $\lambda$ determines the weight of the penalty for constraint violation.

\subsection{Diffused Cost}

When the objective function is evolved according to the diffusion equation (\ref{eq:diffusion}), the diffused objective has a closed form expression. Specifically, it is obtained by the convolution of the original objective with the Gaussian kernel. This can be more formally expressed as the following. Arrange all optimization variables into a long vector $\boldsymbol{w}$, i.e. $\boldsymbol{w} \triangleq \vecc(\boldsymbol{a},\boldsymbol{b}, \boldsymbol{m}_0, \boldsymbol{U},\boldsymbol{V}, \boldsymbol{W}, \boldsymbol{M}, \boldsymbol{N})$. Hence, the cost function can be denoted by $f(\boldsymbol{w})$. The diffused cost function $g$ is obtained by:

\begin{equation}
\label{eq:diffused_cost}
g(\boldsymbol{w}; \sigma) \triangleq [f \star k_\sigma](\boldsymbol{w}) \,.
\end{equation}

After computing this convolution, the variables in $\boldsymbol{w}$ can be replaced by their original names according to the arrangements made in $\boldsymbol{w} \triangleq \vecc(\boldsymbol{a},\boldsymbol{b}, \boldsymbol{m}_0, \boldsymbol{U},\boldsymbol{V}, \boldsymbol{W}, \boldsymbol{M}, \boldsymbol{N})$.

Denote the diffused form of the activation function $h$ by $\tilde{h}_\sigma$, that is $\tilde{h}_\sigma(x) \triangleq [h \star k_\sigma](x)$. Similarly, define $\widetilde{h^2_\sigma}(x) \triangleq [h^2 \star k_\sigma](x)$. The diffused cost w.r.t. optimization variables has the following closed form (see Appendix \ref{sec:diffused_cost}):

\begin{eqnarray}
\sum_{s=1}^S \Bigg( \sum_{t=1}^T & & \| \widetilde{h_\sigma}(\boldsymbol{n}_{s,t})-\boldsymbol{y}_{s,t} \|^2 + \| \sqrt{\widetilde{h_\sigma^2}}(\boldsymbol{n}_{s,t}) \|^2 - \| \widetilde{h_\sigma}(\boldsymbol{n}_{s,t} ) \|^2 \nonumber \\
&+& \lambda \big( \| \boldsymbol{W} \, \widetilde{h_\sigma}(\boldsymbol{m}_{s,t} ) + \boldsymbol{b} - \boldsymbol{n}_{s,t} \|^2 \,+\, \| \boldsymbol{U} \boldsymbol{x}_{s,t} + \boldsymbol{V} \widetilde{h_\sigma}(\boldsymbol{m}_{s,t-1}) \, + \boldsymbol{a} - \boldsymbol{m}_{s,t} \|^2 \nonumber \\
& & \quad + \| \boldsymbol{W} \, \diag(\sqrt{\widetilde{h_\sigma^2}}(\boldsymbol{m}_{s,t} )) \|_F^2 - \| \boldsymbol{W} \, \diag(\widetilde{h_\sigma}(\boldsymbol{m}_{s,t} )) \|_F^2 + \sigma^2 Y \,  \| \widetilde{h_\sigma}(\boldsymbol{m}_{s,t} )\|^2 \big) \nonumber \\
+ \lambda \sum_{t=0}^{T-1} & & \| \boldsymbol{V} \, \diag(\sqrt{\widetilde{h_\sigma^2}}(\boldsymbol{m}_{s,t} )) \|_F^2 - \| \boldsymbol{V} \, \diag(\widetilde{h_\sigma}(\boldsymbol{m}_{s,t} )) \|_F^2 + \sigma^2 H \, \| \widetilde{h_\sigma}(\boldsymbol{m}_{s,t})\|^2 \Bigg)\,. \nonumber
\end{eqnarray}

Here $\| \,.\,\|_F$ denotes the Frobenius norm of a matrix.

\subsection{Approximate Diffused Cost}

Ideal solution requires $S \times T$ auxiliary variables for $\boldsymbol{n}_{s,t}$ and $\boldsymbol{m}_{s,t}$. This is not practical as often $S$ is large. Thus, we resort to an \emc{approximate} formulation here. Instead of solving for the optimal $\boldsymbol{n}_{s,t}$ and $\boldsymbol{m}_{s,t}$, we fix them as below,

\begin{equation}
\label{eq:m_n_def}
\boldsymbol{n}_{s,t} \triangleq \boldsymbol{W} \, \widetilde{h_\sigma}(\boldsymbol{m}_{s,t} ) + \boldsymbol{b} \quad,\quad \boldsymbol{m}_{s,t} \triangleq \boldsymbol{U} \boldsymbol{x}_{s,t} + \boldsymbol{V} \widetilde{h_\sigma}(\boldsymbol{m}_{s,t-1}) \, + \boldsymbol{a} \,.
\end{equation}

This allows us to \emc{drop} $\boldsymbol{n}_{s,t}$ and $\boldsymbol{m}_{s,t}$ from the optimization variables. We computing the gradient, however, derivatives involving $\boldsymbol{n}_{s,t}$ and $\boldsymbol{m}_{s,t}$ must be handled carefully to recognize the dependency stated in (\ref{eq:m_n_def}). The simplified optimization problem is as below,

\begin{eqnarray}
\min_{\boldsymbol{a},\boldsymbol{b}, \boldsymbol{m}_0, \boldsymbol{U},\boldsymbol{V}, \boldsymbol{W}} & & \nonumber \\ \sum_{s=1}^S \Bigg( \sum_{t=1}^T & & \| \widetilde{h_\sigma}(\boldsymbol{n}_{s,t})-\boldsymbol{y}_{s,t} \|^2 + \| \sqrt{\widetilde{h_\sigma^2}}(\boldsymbol{n}_{s,t}) \|^2 - \| \widetilde{h_\sigma}(\boldsymbol{n}_{s,t} ) \|^2 \nonumber \\
&+& \lambda \big( \| \boldsymbol{W} \, \diag(\sqrt{\widetilde{h_\sigma^2}}(\boldsymbol{m}_{s,t} )) \|_F^2 - \| \boldsymbol{W} \, \diag(\widetilde{h_\sigma}(\boldsymbol{m}_{s,t} )) \|_F^2 + \sigma^2 Y \,  \| \widetilde{h_\sigma}(\boldsymbol{m}_{s,t} )\|^2 \big) \nonumber \\
+ \lambda \sum_{t=0}^{T-1} & & \| \boldsymbol{V} \, \diag(\sqrt{\widetilde{h_\sigma^2}}(\boldsymbol{m}_{s,t} )) \|_F^2 - \| \boldsymbol{V} \, \diag(\widetilde{h_\sigma}(\boldsymbol{m}_{s,t} )) \|_F^2 + \sigma^2 H \, \| \widetilde{h_\sigma}(\boldsymbol{m}_{s,t})\|^2 \Bigg) \nonumber \\
\mbox{s.t. }& & \boldsymbol{n}_{s,t} \triangleq \boldsymbol{W} \, \widetilde{h_\sigma}(\boldsymbol{m}_{s,t} ) + \boldsymbol{b} \quad,\quad \boldsymbol{m}_{s,t} \triangleq \boldsymbol{U} \boldsymbol{x}_{s,t} + \boldsymbol{V} \widetilde{h_\sigma}(\boldsymbol{m}_{s,t-1}) \, + \boldsymbol{a} \,. \nonumber
\end{eqnarray}
The gradient of this cost w.r.t. learning parameters are provided in Appendix \ref{sec:gradient}.

\section{Properties of Diffused Cost}

The optimization problem that arises from training a deep network is often challenging. Therefore, local optimization methods (e.g., SGD) are used with a combination of some helping techniques. Although these techniques seem disconnected from each other, some of them emerge automatically from the diffused cost function. Therefore, these techniques might be \emc{unified} under one simple theory. These methods and their connection to the diffused cost are discussed in the following.

\subsection{Careful Initialization}

Local optimization methods are generally sensitive to initialization when it comes to nonconvex cost functions. Deep learning is not an exception \cite{Init13}; a recent study shows that the performance of deep networks and recurrent networks critically depends on initialization \cite{SafranS15}. In contrast, the diffusion algorithm is deterministic and almost independent of initialization\footnote{Path following process could be sensitive to initialization when it reaches a saddle point. Due to instability of saddle points, the direction the algorithm takes could be affected even by small perturbations. Thus, different initializations may end up reaching different solutions. However, these saddle points often occur due to the symmetry in the problem (either the original or the diffused) and the chosen direction does not affect the quality of the solution. This contrasts to gradient descent on a nonconvex objective, where depending on initialization, very solutions of different quality might be reached.} for two reasons. First, after enough smoothing the cost function becomes unimodal, and in case of convexity, will have one global minimum. In fact, the minimizer of the heavily smoothed function coincides with its center mass \cite{mobahi2012phd}. Thus, diffusion provides an interesting deterministic initialization. Second, the update rules are completely deterministic (unless one chooses to use SGD instead of GD for local optimization in Algorithm \ref{alg:alg_goal}) and no notion of randomness is involved in the updates.

\subsection{Annealed Learning Rate}

Each iteration of the gradient descent essentially sees the first order Taylor expansion of the cost function $g(\boldsymbol{x})$ at the current estimate of the solution point $\boldsymbol{x}_0$. The linear approximation has good accuracy only within a small neighborhood of $\boldsymbol{x}_0$, say of radius $\rho$. Enforcing accuracy by the constraint $\| \boldsymbol{x} - \boldsymbol{x}_0 \| \leq \rho$, we arrive at the following problem,

\begin{equation}
\min_{\boldsymbol{x}} g(\boldsymbol{x}_0) + (\boldsymbol{x} - \boldsymbol{x}_0)^T \nabla g(\boldsymbol{x}_0) \quad \quad {s.t.} \quad \| \boldsymbol{x} - \boldsymbol{x}_0 \| \leq \rho \,.
\end{equation}

Using Lagrange multipliers method, the solution of this optimization turns out to be ${\boldsymbol{x}}^* =\boldsymbol{x}_0-\rho \frac{\nabla g(\boldsymbol{x}_0)}{ \|\nabla g(\boldsymbol{x}_0) \|} $.

The radius $\rho$ could be chosen intelligently, e.g., by restricting the tolerated  amount of linearization error. Specifically, in order to ensure $\forall \boldsymbol{x} \,;\, \| \boldsymbol{x} - \boldsymbol{x}_0 \| \leq \rho  \Rightarrow | g(\boldsymbol{x}_0) + (\boldsymbol{x} - \boldsymbol{x}_0)^T \nabla g(\boldsymbol{x}_0) - g(\boldsymbol{x})|\leq \epsilon$, we can choose $\rho = \sqrt{\frac{\epsilon}{c_f}} \sigma $ (see Appendix \ref{sec:linear} for proof). Here $c_f$ is some number satisfying $c_f \geq \frac{1}{2\pi}\sum_{j,k} \| \frac{d^2 f}{d x_j\,  d x_k}\|_{\frac{n}{2}}$, which obviously exists when the norm is bounded.

Putting the pieces together, the solution of the linearized problem can be expressed as ${\boldsymbol{x}}^* =\boldsymbol{x}_0- \eta \, \sigma \, \frac{\nabla g(\boldsymbol{x}_0)}{ \|\nabla g(\boldsymbol{x}_0) \|} $, where $\eta \triangleq \sqrt{\frac{\epsilon}{c_f}}$ is a constant. This is essentially a gradient descent update with a specific choice of the step size. Since $\sigma$ decays toward zero within the continuation loop, the step size (also called learning rate) anneals form an initially large value to eventually a small value.  

\subsection{Noise Injection}
\label{sec:noise_inject}

Injection of random noise into the training process can lead to more stable solutions. This is often crucial in order to obtain satisfactory generalization in deep learning. The well known \emc{dropout} is a specific way of noise injection: in each iteration, it eliminates a random subset of nodes throughout the learning \cite{dropout}. The stochasticity in SGD is another relevant example. It is known that SGD achieves better generalization compared to a full batch gradient descent. More recently, it has been shown that adding Gaussian noise to the computed gradient can significantly improve learning for very deep networks \cite{GradientNoise}. Although these schemes differ in details, e.g., the distribution of the noise or how it is applied to the learning process, they share the same idea of noise injection in learning.

It turns out that the diffused cost function also has this property. In order to see that, recall the definition of the diffused cost function from (\ref{eq:diffused_cost}):

\begin{equation}
g(\boldsymbol{w};\sigma) \triangleq [f \star k_\sigma](\boldsymbol{w}) =\int_{\mathcal{W}} f(\boldsymbol{w}-\boldsymbol{t}) k_\sigma(\boldsymbol{t}) \, d \boldsymbol{t}
\end{equation}

Thus, the gradient at a point $\boldsymbol{w}_0$ has the following form.
\begin{eqnarray}
\label{eq:diffused_gradient}
\nabla g(\boldsymbol{w}_0;\sigma) &=& \int_{\mathcal{W}} \nabla f(\boldsymbol{w}_0-\boldsymbol{t}) k_\sigma(\boldsymbol{t}) \, d \boldsymbol{t} \\
\label{eq:diffused_gradient_sample}
&\approx& \frac{1}{J} \sum_{j=1}^J \nabla f(\boldsymbol{w}_0-\boldsymbol{t}_j) \quad,\quad \boldsymbol{t}_j \sim \mathcal{N}(\boldsymbol{0},\sigma^2 \boldsymbol{I}) \,.
\end{eqnarray}

This means if we were to approximate the gradient of the diffused cost by MCMC method, it would average over a number of \emc{noisified} gradients. Specifically, the noise would be \emc{additive} w.r.t. the weights of the network and it would have a \emc{normal distribution} with zero mean and variance of $\sigma^2$. The noise injection of (\ref{eq:diffused_gradient_sample}) has also been used by \cite{fuzz} via numerical sampling exactly as in (\ref{eq:diffused_gradient_sample}). From a higher level perspective, this noise injection has some similarity to SGD; the latter also averages (over multiple epochs) the effect of noisified gradients.

A key advantage of using the diffusion framework for noise injection, however, is that the expected noisified gradient (the integral in (\ref{eq:diffused_gradient})) has a \emc{closed form} expression, while the other schemes are mainly \emc{sampling} based. This leads to a huge computational gain for the diffusion method: while other methods would need a lot of sampling iterations in order to reach a reasonable approximation to the expected noisified gradient (and the number of these samples could grow exponentially in the number of weights), the diffusion method achieves this with almost no computational effort and without any sampling.

\subsection{Layerwise Pretraining}

We argue that when $\sigma$ is large, the network only focuses on short range dependencies, and as $\sigma$ shrinks toward zero, longer range dependencies are gradually learned. In order to see why this happens, let's for example inspect the partial gradient $\nabla_{\boldsymbol{a}} \, {g}$, which has the form $\sum_{t=1}^T \boldsymbol{r}_t \, \boldsymbol{M}_t$ (see Appendix \ref{sec:gradient} for derivations and the definition of $\boldsymbol{r}_t$), where $\boldsymbol{M}_t \triangleq  \boldsymbol{I} + \boldsymbol{V} \diag \big( {\widetilde{h}}^\prime (\boldsymbol{m}_{t-1}) \big) \boldsymbol{M}_{t-1}$ and $\boldsymbol{M}_1 \triangleq \boldsymbol{I}$. Resolving the recursion in $\boldsymbol{M}_t$ leads to,

\begin{equation}
\boldsymbol{M}_t = \boldsymbol{I} + \boldsymbol{V} \diag \big( {\widetilde{h}}_\sigma^\prime (\boldsymbol{m}_{t-1}) \big) + \boldsymbol{V} {\widetilde{h}}_\sigma^\prime (\boldsymbol{m}_{t-1}) \, \boldsymbol{V} {\widetilde{h}}_\sigma^\prime (\boldsymbol{m}_{t-2})+ \dots \,. \nonumber
\end{equation}

When $\sigma \rightarrow \infty$, all the sigmoid-like activation functions listed in (\ref{tab:Table}) become flat and their gradient vanishes ${\widetilde{h}}_\sigma^\prime \rightarrow 0$. This implies that by choosing $\sigma$ large enough, one can find a small enough $\epsilon$ that satisfies $\|\diag({\widetilde{h}}_\sigma^\prime)\| \leq \epsilon$. Since the contribution of each term in the above sum will be at most equal to its matrix norm, we can derive,

\begin{equation}
\|\boldsymbol{M}_t\| \leq \|\boldsymbol{I}\| + \epsilon \|\boldsymbol{V}\| + (\epsilon \|\boldsymbol{V}\|)^2 + (\epsilon \|\boldsymbol{V}\|)^3 +  \dots \,. \nonumber
\end{equation}

when $\sigma$ is very large, and thus $\epsilon$ is very small, we can ignore all the terms involving $\epsilon$, which leaves us with $\boldsymbol{M}_t\approx \boldsymbol{I}$. As we gradually reduce $\sigma$, and thus increase $\epsilon$, we can reconsider terms involving smaller exponents, while the higher order terms still remain negligible. By gradually decreasing $\sigma$, $\boldsymbol{M}_t$ can be approximated by $\boldsymbol{I}$, then, $\boldsymbol{I} + \boldsymbol{V} \diag \big( {\widetilde{h}}_\sigma^\prime (\boldsymbol{m}_{t-1}) \big) $, then $\boldsymbol{I} + \boldsymbol{V} \diag \big( {\widetilde{h}}_\sigma^\prime (\boldsymbol{m}_{t-1}) \big) + \boldsymbol{V} {\widetilde{h}}_\sigma^\prime (\boldsymbol{m}_{t-1}) \, \boldsymbol{V} {\widetilde{h}}_\sigma^\prime (\boldsymbol{m}_{t-2})$ and so on. 

This is conceptually very similar to layerwise pretraining \cite{Bengio06}, as the learning in each layer starts from considering only its immediate previous layer and then gradually switches to the full consideration by considering larger and larger number of previous layers. For example, $M_t$ at the first layer (i.e., when $t=1$) first considers contribution from itself only, then gradually introduces contribution from the second layer, and then from the third layer and so on.

\section{Choice of the Activation Function}

In order to implement the method, we need to obtain the explicit expressions of $\widetilde{h}_\sigma$ and $\widetilde{h^2}_\sigma$ for a given activation function $h$. For example, suppose we set $h(x)=\erf(a x)$, where $a$ is a parameter that determines the sharpness of the activation function. Note that $\lim_{a \rightarrow \infty} \erf(ax) = \sign(x)$ and $\erf(\frac{\sqrt{\pi}}{2} x) \approx \tanh(x)$. The form of $\widetilde{h}_\sigma$ can be already looked up from Table \ref{tab:Table}, which is repeated below,

\begin{equation}
\tilde{h}(x) = \erf(\frac{a x}{\sqrt{1+2 (a \sigma)^2}}) \,.
\end{equation}

\begin{figure}
\centering \includegraphics[width=.4\textwidth,height=.3\textwidth]{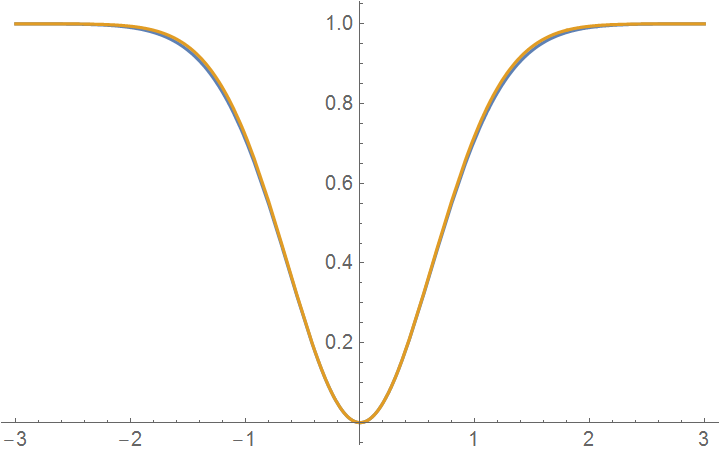}
\caption{Blue and brown curves respectively plot $\erf^2(x)$ and $1-e^{-\frac{4}{\pi} x^2}$. Due to the strong overlap, the blue curve is barely visible.}
\label{fig:erf2}
\end{figure}

In the following, we only focus on $\widetilde{h^2}_\sigma$. Unfortunately, $\widetilde{h^2}_\sigma(x)$ lacks a closed form expression. However, observe that $\erf^2(x) \approx 1-e^{-\frac{4}{\pi} x^2}$. This approximation has a reasonably good accuracy as shown in Figure \ref{fig:erf2}. Using this approximation, it follows that $[\erf^2(a \,\Box\, ) \star k_\sigma] (x) \approx [1-e^{-\frac{4}{\pi} (a \,\Box\, )^2}  k_\sigma] (x)$. 

\begin{eqnarray}
\widetilde{h^2}(x) &\triangleq& [\erf^2(a \,\Box\, ) \star k_\sigma] (x) \nonumber \\
&\approx& [1-e^{-\frac{4}{\pi} (a \,\Box\, )^2}  k_\sigma] (x) \nonumber \\
&=& 1- \frac{\sqrt{\pi} \,\, e^{-\frac{4 a^2 x^2}{\pi + 8 a^2 \sigma^2}}}{\sqrt{\pi + 8 a^2 \sigma^2}} \nonumber \,.
\end{eqnarray}

\section{Preliminary Results}

Here we present a comparison between SGD and the proposed diffusion framework. The hyperparameters in both methods are carefully searched to ensure a fair comparison. We use $\erf$ as the activation function. The task is to learn adding two numbers, and is adapted from \cite{RNNHard}. The network consists of has 10 hidden units, and it has two inputs and one output. One of the input units reads a sequence of 10 real numbers, and the other a sequence of 10 binary numbers. The binary numbers are zero everywhere except two random locations. The task is to add the values from the first sequence, at the two locations marked by the second sequence.

\begin{figure}
\begin{center}
\includegraphics[width=.8\textwidth]{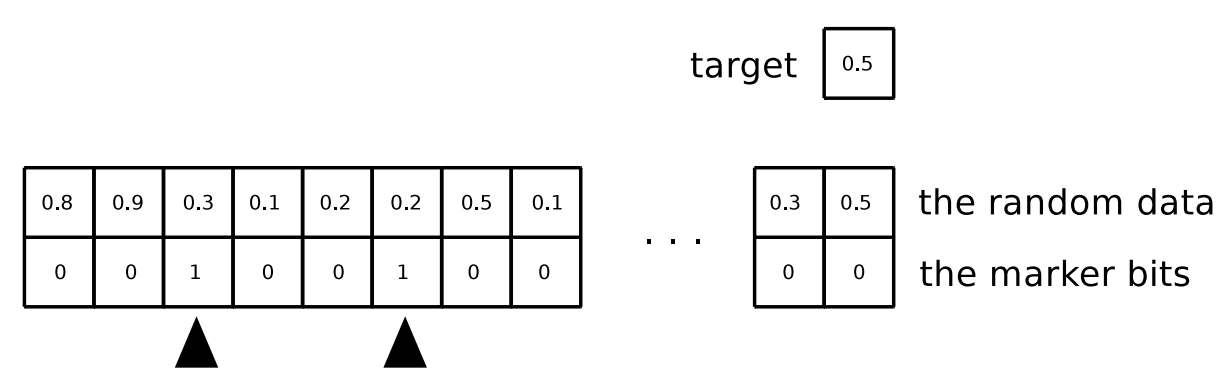}
\caption{Learning to add by RNNs. Figure adapted with permission from \cite{RNNHard}.}
\end{center}
\end{figure}

We trained the network by 1000 sequences, and generalization is computed from a test set of 100 sequences. The result is shown in the plots. The horizontal axis shows the \emc{generalization error}, and the vertical axis shows how many \emc{epochs} it takes to reach that generalization error. For example, with 50 batches of size 50 samples, in order to reach around error of 0.02, SGD (blue) needs about 90 epochs, while diffusion methods (red) needs about 20 epochs. 

\begin{figure}
\noindent\begin{center}
\includegraphics[width=.45\textwidth]{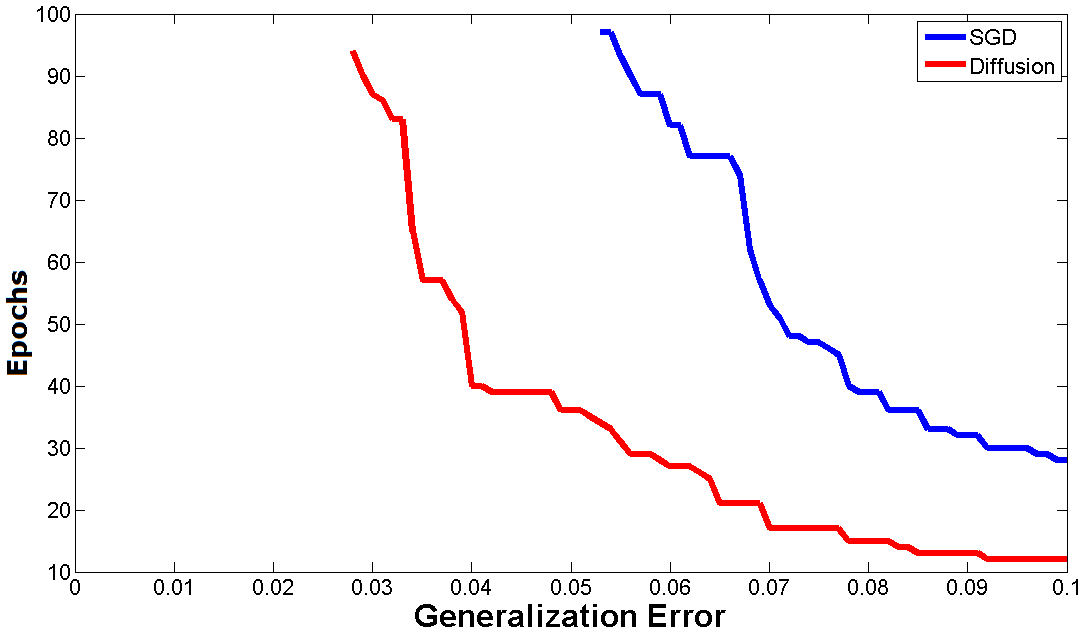}
\includegraphics[width=.45\textwidth]{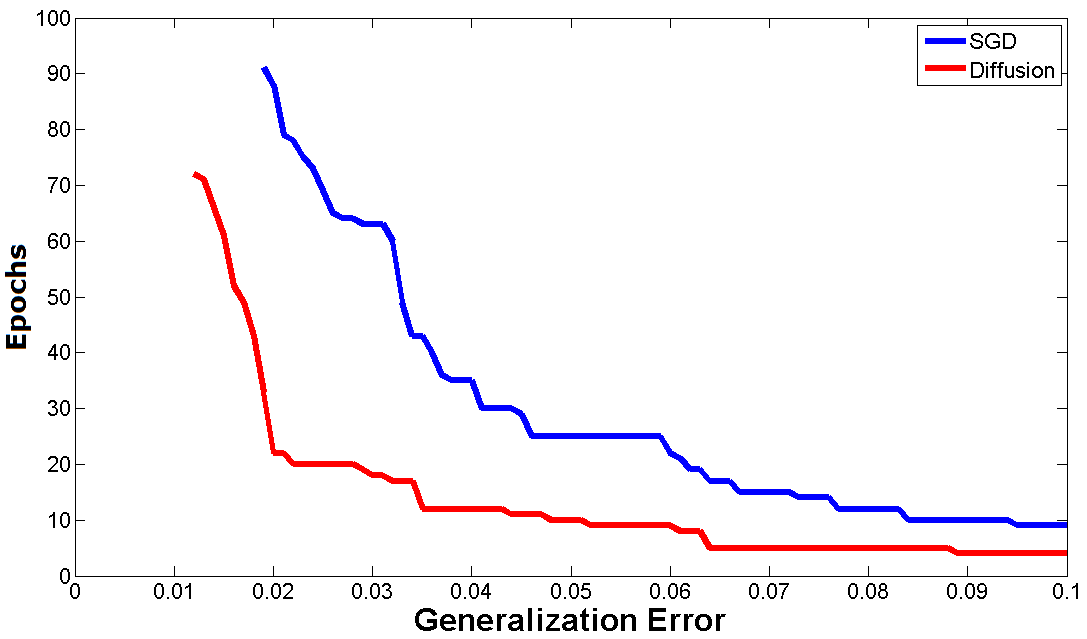}
\end{center}
\caption{Experiments with mini batches of size 10 (left) and 50 (right).}
\end{figure}

\section{Related Works \& Future Directions}

This work specifically studies the use of the diffusion equation for optimizing the objective function in deep learning. However, there is a growing number of techniques by others that propose new algorithms for deep learning. Using tensor decomposition techniques, \cite{anima} offers new algorithms for deep learning with performance guarantee. A conceptually similar algorithm to ours is provided in \cite{hazan}. However, instead of computing the convolution analytically, the latter work relies on numerical sampling. It guarantees reaching the global minimum at a proved rate for certain objective functions.

This work relies on smoothing the objective function by convolving it with the Gaussian kernel. We have previously shown that this particular form of smoothing is optimal in a certain sense, by relating Gaussian convolution to a relaxation of the convex envelope. Although connection to the convex envelope is meaningful in the context of \emc{nonconvex} objective functions, there are side benefits in smoothing even when the objective function is convex. For example, smoothing a nonsmooth \emc{convex} objective function by convolution can improve the convergence rate of stochastic optimization algorithms \cite{Duchi}.

As discussed in Section \ref{sec:noise_inject}, smoothing can be considered as means to inject noise into the training process. The idea of noise injection is already used in methods such as SGD or dropout \cite{dropout} in order to improve learning. The key advantage of our framework for noise injection, however, is that the noise injection can be achieved in closed form and without need of sampling. In order words, we can compute the effect of infinitely many noisified objective functions in closed form. This is similar to the idea of Marginalized Denoising Autoencoders (mDA) \cite{feisha}, where the effect of infinitely many nosified inputs is marginalized to obtain a closed form expression. However, mDA limits the form of the injected noise. Specifically, the marginalized effect is only computable in a \emc{linear} reconstruction setup (nonlinearity is applied only after computation of the marginalized reconstruction). In addition, mDA performs noise injection layer by layer in a greedy fashion. In contrast, our framework is able to compute closed form expression for the entire deep network and allowing full nonconvexity of the associated optimization, up to reasonable approximation.

\begin{figure}
\centering
\includegraphics[width=.6\textwidth]{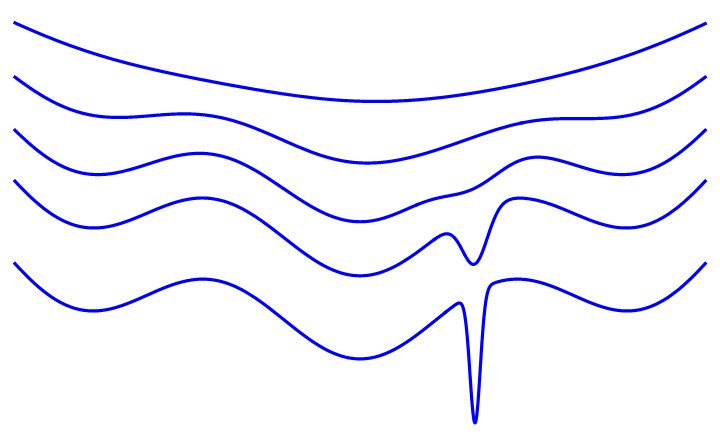}
\caption{Starting from the original function at the bottom, moving upward the plots correspond to more aggressive smoothing (i.e. larger $\sigma)$. The original function has three wide minima, and a narrow global minimum. Following the path of the minimizer from top to the bottom, it is obvious that the process misses the narrow global minimum and reaches one of the wider minima. However, among the three wide minima, it finds the lowest one.}
\label{fig:unstable}
\end{figure}

Diffusion equation provides an approximate evolution toward the convex envelope. Consequently, it is not perfect: if global minimum is very narrow, diffusion can miss that minima in favor of a wider minimum whose value is slightly larger than the narrow global minimum (see Figure \ref{fig:unstable}). This may seem a disadvantage at the first glance. However, the wider minima are in fact more stable\footnote{By a stable minimum we mean that a small perturbation of the equilibrium resides in the basin of attraction of the same equilibrium. This is not true if the minimum is too narrow; slight perturbation may put the gradient decent into a different basin of attraction.}, which could be more desired in practice, e.g. generalizing better. In fact, a recent analysis has shown that SGD attains better generalization when the objective function is \emc{smoother} \cite{recht}. Note that in our framework, initializing the algorithm with larger $\sigma$ automatically provides a smoother surrogate cost function where unstable minima disappear. Thus, it is more likely to remain in the basin of attraction of the stable minima. A thorough investigation of how smoothing the cost function in the diffusion setting may improve the generalization performance is a direction for future research.

A closely related work to ours is Annealed Gradient Descent \cite{annealedPan}, where the objective landscape is also initially approximated by a smoother function and is gradually transformed to the original one. However, the unlike this work where Gaussian smoothing is theoretically motivated for nonconvex optimization  \cite{MobahiEMMCVPR}, in \cite{annealedPan} coarse-to-fine approximation of the objective function is based on heuristically motivated procedure. More precisely, the latter uses vector quantization methods in order to generate a code book by which the coarse cost function is approximated. Another difference between these two works is that the representation of the smoothed function in our framework is simpler, as we directly obtain a closed form expression of the objective function. that is a simpler setup than approximation by codebook generation. 

Very recently, the diffusion process has been proposed for learning difficult probability distributions \cite{ganguli}. In forward time diffusion, the method converts any complex distribution into a simple distribution, e.g., Gaussian. It then learns the reverse-time of this diffusion process to define a generative
model distribution. By sampling from such trained model, the authors have achieved inpainting of missing regions in natural images.

\section{Acknowledgment}

This research is partially funded by Shell Research. Hossein Mobahi is thankful to John W. Fisher, William T. Freeman, Yann LeCun, and Yoshua Bengio for comments and discussions and to Peter Bartlett and Fei Sha for suggesting connections to \cite{Duchi,feisha}. Hossein Mobahi is grateful to Geoffrey Hinton, Marc'Aurelio Ranzato, and Philip Bachman for comments and Kate Saenko for discussions in earlier phase of this work.

\bibliography{fast_rnn_training}

\begin{thebibliography}{}

\bibitem[Ackley, 1987]{ackley}
Ackley, D. (1987).
\newblock {\em A Connectionist Machine for Genetic Hillclimbing}, volume SECS28
  of {\em The Kluwer International Series in Engineering and Computer Science.}
\newblock Kluwer Academic Publishers, Boston.

\bibitem[Bachman et~al., 2014]{fuzz}
Bachman, P., Alsharif, O., and Precup, D. (2014).
\newblock Learning with pseudo-ensembles.
\newblock In {\em Advances in Neural Information Processing Systems 27}.

\bibitem[Bahdanau et~al., 2014]{BahdanauCB14}
Bahdanau, D., Cho, K., and Bengio, Y. (2014).
\newblock Neural machine translation by jointly learning to align and
  translate.
\newblock {\em CoRR}, abs/1409.0473.

\bibitem[Bengio et~al., 2007]{Bengio06}
Bengio, Y., Lamblin, P., Popovici, D., and Larochelle, H. (2007).
\newblock Greedy layer-wise training of deep networks.
\newblock In {\em Advances in Neural Information Processing Systems 19}.

\bibitem[Bengio et~al., 2009]{Beng09Cur}
Bengio, Y., Louradour, J., Collobert, R., and Weston, J. (2009).
\newblock Curriculum learning.
\newblock In {\em {ICML}}.

\bibitem[Chen et~al., 2014]{feisha}
Chen, M., Weinberger, K.~Q., Sha, F., and Bengio, Y. (2014).
\newblock Marginalized denoising auto-encoders for nonlinear representations.
\newblock In {\em Proceedings of the 31st International Conference on Machine
  Learning (ICML-14)}, pages 1476--1484.

\bibitem[Dauphin et~al., 2014]{Dauphin}
Dauphin, Y.~N., Pascanu, R., Gulcehre, C., Cho, K., Ganguli, S., and Bengio, Y.
  (2014).
\newblock Identifying and attacking the saddle point problem in
  high-dimensional non-convex optimization.
\newblock pages 2933--2941.

\bibitem[Duchi et~al., 2012]{Duchi}
Duchi, J.~C., Bartlett, P.~L., and Wainwright, M.~J. (2012).
\newblock Randomized smoothing for stochastic optimization.
\newblock {\em SIAM Journal on Optimization}, 22(2):674--701.

\bibitem[Hardt et~al., 2015]{recht}
Hardt, M., Recht, B., and Singer, Y. (2015).
\newblock Train faster, generalize better: Stability of stochastic gradient
  descent.
\newblock {\em CoRR}, abs/1509.01240.

\bibitem[Hazan et~al., 2015]{hazan}
Hazan, E., Levy, K.~Y., and Shalev{-}Shwartz, S. (2015).
\newblock On graduated optimization for stochastic non-convex problems.
\newblock {\em CoRR}, abs/1503.03712.

\bibitem[Hinton et~al., 2012a]{HintonSpeech}
Hinton, G.~E., Deng, L., Yu, D., Dahl, G.~E., Mohamed, A., Jaitly, N., Senior,
  A., Vanhoucke, V., Nguyen, P., Sainath, T.~N., and Kingsbury, B. (2012a).
\newblock Deep neural networks for acoustic modeling in speech recognition: The
  shared views of four research groups.
\newblock {\em IEEE Signal Process. Mag.}, 29(6):82--97.

\bibitem[Hinton et~al., 2012b]{dropout}
Hinton, G.~E., Srivastava, N., Krizhevsky, A., Sutskever, I., and
  Salakhutdinov, R. (2012b).
\newblock Improving neural networks by preventing co-adaptation of feature
  detectors.
\newblock {\em CoRR}, abs/1207.0580.

\bibitem[Janzamin et~al., 2015]{anima}
Janzamin, M., Sedghi, H., and Anandkumar, A. (2015).
\newblock Generalization bounds for neural networks through tensor
  factorization.
\newblock {\em CoRR}, abs/1506.08473.

\bibitem[Krizhevsky et~al., 2012]{AlexNet}
Krizhevsky, A., Sutskever, I., and Hinton, G.~E. (2012).
\newblock Imagenet classification with deep convolutional neural networks.
\newblock In {\em Advances in Neural Information Processing Systems 25}.

\bibitem[Martens and Sutskever, 2011]{RNNHard}
Martens, J. and Sutskever, I. (2011).
\newblock Learning recurrent neural networks with hessian-free optimization.
\newblock In {\em ICML}, pages 1033--1040. Omnipress.

\bibitem[Mobahi, 2012]{mobahi2012phd}
Mobahi, H. (2012).
\newblock {\em Optimization by Gaussian Smoothing with Application to Geometric
  Alignment}.
\newblock PhD thesis, University of Illinois at Urbana Champaign.

\bibitem[Mobahi, 2016]{mobahiclosed}
Mobahi, H. (2016).
\newblock Closed form for some gaussian convolutions.
\newblock {\em CoRR}.

\bibitem[Mobahi and Fisher~III, 2015]{MobahiEMMCVPR}
Mobahi, H. and Fisher~III, J.~W. (2015).
\newblock {On the Link Between Gaussian Homotopy Continuation and Convex
  Envelopes}.

\bibitem[Neelakantan et~al., 2015]{GradientNoise}
Neelakantan, A., Vilnis, L., Le, Q.~V., Sutskever, I., Kaiser, L., Kurach, K.,
  and Martens, J. (2015).
\newblock Adding gradient noise improves learning for very deep networks.
\newblock {\em CoRR}, abs/1511.06807.

\bibitem[Pan and Jiang, 2015]{annealedPan}
Pan, H. and Jiang, H. (2015).
\newblock Annealed gradient descent for deep learning.
\newblock In {\em Proc. of 31th Conference on Uncertainty in Artificial
  Intelligence (UAI 2015)}.

\bibitem[Pascanu et~al., 2014]{PascanuDGB14}
Pascanu, R., Dauphin, Y.~N., Ganguli, S., and Bengio, Y. (2014).
\newblock On the saddle point problem for non-convex optimization.
\newblock {\em CoRR}, abs/1405.4604.

\bibitem[Safran and Shamir, 2015]{SafranS15}
Safran, I. and Shamir, O. (2015).
\newblock On the quality of the initial basin in overspecified neural networks.
\newblock {\em CoRR}, abs/1511.04210.

\bibitem[Sohl{-}Dickstein et~al., 2015]{ganguli}
Sohl{-}Dickstein, J., Weiss, E.~A., Maheswaranathan, N., and Ganguli, S.
  (2015).
\newblock Deep unsupervised learning using nonequilibrium thermodynamics.
\newblock {\em CoRR}, abs/1503.03585.

\bibitem[Sutskever et~al., 2013]{Init13}
Sutskever, I., Martens, J., Dahl, G., and Hinton, G. (2013).
\newblock On the importance of initialization and momentum in deep learning.
\newblock In {\em Proceedings of the 30th International Conference on Machine
  Learning (ICML-13)}.

\bibitem[Sutskever et~al., 2014]{Sut}
Sutskever, I., Vinyals, O., and Le, Q.~V. (2014).
\newblock Sequence to sequence learning with neural networks.
\newblock In Ghahramani, Z., Welling, M., Cortes, C., Lawrence, N., and
  Weinberger, K., editors, {\em Advances in Neural Information Processing
  Systems 27}, pages 3104--3112.

\bibitem[Vese, 1999]{Vese}
Vese, L. (1999).
\newblock {A method to convexify functions via curve evolution.}
\newblock {\em Commun. Partial Differ. Equations}, 24(9-10):1573--1591.

\bibitem[Widder, 1975]{Widder75}
Widder, D.~V. (1975).
\newblock {\em The Heat Equation}.
\newblock Academic Press.

\bibitem[Zhang et~al., 2015]{jordan}
Zhang, Y., Lee, J.~D., and Jordan, M.~I. (2015).
\newblock $\ell_1$-regularized neural networks are improperly learnable in
  polynomial time.
\newblock {\em CoRR}, abs/1510.03528.

\end{thebibliography}
\bibliographystyle{apalike}

\newpage~\newpage~

\noindent {\bf \LARGE Appendices}
\bigskip

\appendix

\section{Diffused RNN Training Cost}
\label{sec:diffused_cost}

Diffusing the cost function w.r.t. $\boldsymbol{a},\boldsymbol{b}, \boldsymbol{U},\boldsymbol{V}, \boldsymbol{W}$ yields\footnote{We use the fact that convolution of $(\boldsymbol{x}^T \boldsymbol{y})^2$ with $k_\sigma(\boldsymbol{x})$ is $(\boldsymbol{x}^T \boldsymbol{y})^2 + \sigma^2 \|\boldsymbol{y}\|^2$.},

\begin{eqnarray}
\sum_{t=1}^T & & \| h(\boldsymbol{n}_t)-\boldsymbol{y}_t \|^2 \\
&+& \lambda \big( \| \boldsymbol{W} \, h(\boldsymbol{m}_t ) + \boldsymbol{b} - \boldsymbol{n}_t \|^2 \,+\, \| \boldsymbol{U} \boldsymbol{x}_t + \boldsymbol{V} h(\boldsymbol{m}_{t-1}) \, + \boldsymbol{a} - \boldsymbol{m}_t \|^2\\
& & \quad + \sigma^2 Y (1+ \| h(\boldsymbol{m}_t )\|^2)  \,+\, \sigma^2 H (1+ \| \boldsymbol{x}_t\|^2 + \| h(\boldsymbol{m}_{t-1})\|^2) \big)\,.
\end{eqnarray}

Smoothing w.r.t. $\boldsymbol{m}_t$ and $\boldsymbol{n}_t$ leads\footnote{We use the identity that convolution of $\| \boldsymbol{A} h (\boldsymbol{x}) + \boldsymbol{b}\|^2$ with $k_\sigma(\boldsymbol{x})$ is equal to $\| \boldsymbol{A} \tilde{h} (\boldsymbol{x}) + \boldsymbol{b}\|^2 +  \| \boldsymbol{A} \, \diag(\sqrt{\widetilde{h^2}}(\boldsymbol{x} )) \|_F^2 - \| \boldsymbol{A} \, \diag(\widetilde{h}(\boldsymbol{x} )) \|_F^2$.} to,

\begin{eqnarray}
\sum_{t=1}^T & & \| \widetilde{h}(\boldsymbol{n}_t)-\boldsymbol{y}_t \|^2 + \| \sqrt{\widetilde{h^2}}(\boldsymbol{n}_t) \|^2 - \| \widetilde{h}(\boldsymbol{n}_t ) \|^2\\
&+& \lambda \big( \| \boldsymbol{W} \, \widetilde{h}(\boldsymbol{m}_t ) + \boldsymbol{b} - \boldsymbol{n}_t \|^2 \,+\, \| \boldsymbol{U} \boldsymbol{x}_t + \boldsymbol{V} \widetilde{h}(\boldsymbol{m}_{t-1}) \, + \boldsymbol{a} - \boldsymbol{m}_t \|^2\\
& & \quad + \sigma^2 Y ( 2 + \| \widetilde{h}(\boldsymbol{m}_t )\|^2)  \,+\, \sigma^2 H (2 + \| \boldsymbol{x}_t\|^2 + \| \widetilde{h}(\boldsymbol{m}_{t-1})\|^2) \\
& & \quad + \| \boldsymbol{W} \, \diag(\sqrt{\widetilde{h^2}}(\boldsymbol{m}_t )) \|_F^2 - \| \boldsymbol{W} \, \diag(\widetilde{h}(\boldsymbol{m}_t )) \|_F^2  \\
& & \quad + \| \boldsymbol{V} \, \diag(\sqrt{\widetilde{h^2}}(\boldsymbol{m}_{t-1} )) \|_F^2 - \| \boldsymbol{V} \, \diag(\widetilde{h}(\boldsymbol{m}_{t-1} )) \|_F^2 \big) \,.
\end{eqnarray}

Discarding constants terms, i.e. those that do not depend on neither of optimization variables $\boldsymbol{a},\boldsymbol{b}, \boldsymbol{U},\boldsymbol{V}, \boldsymbol{W}, \boldsymbol{M} , \boldsymbol{N}$, simplifies the diffused cost to the following,

\begin{eqnarray}
\sum_{t=1}^T & & \| \widetilde{h}(\boldsymbol{n}_t)-\boldsymbol{y}_t \|^2 + \| \sqrt{\widetilde{h^2}}(\boldsymbol{n}_t) \|^2 - \| \widetilde{h}(\boldsymbol{n}_t ) \|^2\\
&+& \lambda \big( \| \boldsymbol{W} \, \widetilde{h}(\boldsymbol{m}_t ) + \boldsymbol{b} - \boldsymbol{n}_t \|^2 \,+\, \| \boldsymbol{U} \boldsymbol{x}_t + \boldsymbol{V} \widetilde{h}(\boldsymbol{m}_{t-1}) \, + \boldsymbol{a} - \boldsymbol{m}_t \|^2\\
& & \quad + \| \boldsymbol{W} \, \diag(\sqrt{\widetilde{h^2}}(\boldsymbol{m}_t )) \|_F^2 - \| \boldsymbol{W} \, \diag(\widetilde{h}(\boldsymbol{m}_t )) \|_F^2 + \sigma^2 Y \,  \| \widetilde{h}(\boldsymbol{m}_t )\|^2 \big) \\
+ \lambda \sum_{t=0}^{T-1} & & \| \boldsymbol{V} \, \diag(\sqrt{\widetilde{h^2}}(\boldsymbol{m}_t )) \|_F^2 - \| \boldsymbol{V} \, \diag(\widetilde{h}(\boldsymbol{m}_t )) \|_F^2 + \sigma^2 H \, \| \widetilde{h}(\boldsymbol{m}_t)\|^2 \,.
\end{eqnarray}

\section{Gradient of Diffused Cost}
\label{sec:gradient}

Below $\odot$ denotes the element-wise product of two matrices.

\begin{eqnarray}
\frac{d g}{d \boldsymbol{b}} &=& \sum_t \frac{\partial \boldsymbol{n}_t}{\partial \boldsymbol{b}} \, \frac{\partial g}{\partial \boldsymbol{n}_t} \\
&=& \sum_t \boldsymbol{I} \Big (2 {\widetilde{h}}^\prime (\boldsymbol{n}_t) \odot (\widetilde{h}(\boldsymbol{n}_t)-\boldsymbol{y}_t) + {\widetilde{h^2}}^\prime(\boldsymbol{n}_t) - 2 {\widetilde{h}}^\prime(\boldsymbol{n}_t ) \odot \widetilde{h}(\boldsymbol{n}_t )  \Big) \\
&=& \sum_t \Big ({\widetilde{h^2}}^\prime(\boldsymbol{n}_t) -2 {\widetilde{h}}^\prime (\boldsymbol{n}_t) \odot \boldsymbol{y}_t \Big) \,.
\end{eqnarray}

\linesep

\begin{eqnarray}
\frac{d g}{d \boldsymbol{W}} &=& \sum_t \frac{\partial g}{\partial \boldsymbol{W}} + \sum_d \frac{\partial g}{\partial n_t^{(d)}} \frac{\partial n_t^{(d)}}{\partial \boldsymbol{W}} \\
&=& 2 \lambda \boldsymbol{W} \diag \Big( \sum_{t=1}^T \big(\widetilde{h^2}(\boldsymbol{m}_t ) \,-\, {\widetilde{h}}^2(\boldsymbol{m}_t ) \big) \Big)\\
& & + \sum_{t=1}^T \Big ({\widetilde{h^2}}^\prime(\boldsymbol{n}_t) -2 {\widetilde{h}}^\prime (\boldsymbol{n}_t) \odot \boldsymbol{y}_t \Big)\, \widetilde{h}(\boldsymbol{m}_t )^T 
\end{eqnarray}

\linesep

\begin{eqnarray}
\boldsymbol{r}_t &\triangleq& \Big ({\widetilde{h^2}}^\prime(\boldsymbol{n}_t) -2 {\widetilde{h}}^\prime (\boldsymbol{n}_t) \odot \boldsymbol{y}_t \Big)^T \, \Big( \boldsymbol{W} \diag({\widetilde{h}}^\prime (\boldsymbol{m}_t))  \Big) \nonumber \\
& & \quad + \lambda \Big( \big( {\widetilde{h^2}}^\prime(\boldsymbol{m}_t) -2 {\widetilde{h}}^\prime(\boldsymbol{m}_t) \odot {\widetilde{h}}(\boldsymbol{m}_t) \big)^T \odot \big(\boldsymbol{1}^T (\boldsymbol{W} \odot \boldsymbol{W}) + \mathbb{I}_{t \neq T}\boldsymbol{1}^T (\boldsymbol{V} \odot \boldsymbol{V}) \big) \nonumber \\
& & \quad\quad\quad  + 2 \sigma^2 (\mathbb{I}_{t \neq T} H + Y) ({\widetilde{h}}^\prime(\boldsymbol{m}_t) \odot {\widetilde{h}}(\boldsymbol{m}_t)  )^T \Big) \,.
\end{eqnarray}

\linesep

\begin{eqnarray}
(\frac{d g}{d \boldsymbol{a}})^T &=& \sum_{t=1}^T (\frac{d g}{d \boldsymbol{m}_t})^T \, \frac{d \boldsymbol{m}_t}{d \boldsymbol{a}} \, \\
&=& \sum_{t=1}^T ((\frac{\partial g}{\partial \boldsymbol{n}_t})^T \, \frac{\partial \boldsymbol{n}_t}{\partial \boldsymbol{m}_t} + (\frac{\partial g}{\partial \boldsymbol{m}_t})^T) \, \frac{d \boldsymbol{m}_t}{d \boldsymbol{a}} \, \\
&=& \sum_{t=1}^T \boldsymbol{r}_t \, \boldsymbol{M}_t \\
\boldsymbol{M}_t &\triangleq& \frac{d \boldsymbol{m}_t}{d \boldsymbol{a}} = \frac{\partial \boldsymbol{m}_t}{\partial \boldsymbol{a}} + \frac{\partial \boldsymbol{m}_t}{\partial \boldsymbol{m}_{t-1}} \boldsymbol{M}_{t-1} = \boldsymbol{I} + \boldsymbol{V} \diag \big( {\widetilde{h}}^\prime (\boldsymbol{m}_{t-1}) \big) \boldsymbol{M}_{t-1}\\
\boldsymbol{M}_1 &\triangleq& \boldsymbol{I} \,.
\end{eqnarray}

\linesep

\begin{eqnarray}
\frac{d g}{d \boldsymbol{V}} &=& \frac{\partial g}{\partial \boldsymbol{V}} + \sum_{t=1}^T \sum_d \frac{d g}{d m_t^{(d)}} \frac{d m_t^{(d)}}{d \boldsymbol{V}}  \\
&=& \frac{\partial g}{\partial \boldsymbol{V}} + \sum_{t=1}^T \sum_d ((\frac{\partial g}{\partial \boldsymbol{n}_t})^T \, \frac{\partial \boldsymbol{n}_t}{\partial \boldsymbol{m}_t} + (\frac{\partial g}{\partial \boldsymbol{m}_t})^T)^{(d)} \frac{d m_t^{(d)}}{d \boldsymbol{V}}  \\
&=& 2 \lambda \boldsymbol{V} \diag \Big( \sum_{t=0}^{T-1} \big(\widetilde{h^2}(\boldsymbol{m}_t ) \,-\, {\widetilde{h}}^2(\boldsymbol{m}_t ) \big) \Big) + \sum_{t=1}^T \sum_d  \boldsymbol{r}_t^{(d)} \, \boldsymbol{M}_t^{(d)}  \\
\boldsymbol{M}_t^{(d)} &\triangleq& \frac{d \boldsymbol{m}_t^{(d)}}{d \boldsymbol{V}}\\
&=& \frac{\partial \boldsymbol{m}_t^{(d)}}{\partial \boldsymbol{V}} + \sum_{d^\prime} \frac{\partial \boldsymbol{m}_t^{(d)}}{\partial \boldsymbol{m}_{t-1}^{(d^\prime)}} \boldsymbol{M}_{t-1}^{(d^\prime)}\\
&=& \mbox{"Zero matrix except d'th row set to $\widetilde{h}^T(\boldsymbol{m}_{t-1})$"} + \sum_{d^\prime} v_{d,d^\prime}  {\widetilde{h}}^\prime (m_{t-1}^{(d^\prime)}) \boldsymbol{M}_{t-1}^{(d^\prime)} \\
\boldsymbol{M}_1^{(d)} &\triangleq& \mbox{"Zero matrix except d'th row set to $\widetilde{h}^T(\boldsymbol{m}_{0})$"} \,.
\end{eqnarray}

\linesep

\begin{eqnarray}
\frac{d g}{d \boldsymbol{U}} &=& \frac{\partial g}{\partial \boldsymbol{U}} + \sum_{t=1}^T \sum_d \frac{d g}{d m_t^{(d)}} \frac{d m_t^{(d)}}{d \boldsymbol{U}}  \\
&=& \frac{\partial g}{\partial \boldsymbol{U}} + \sum_{t=1}^T \sum_d ((\frac{\partial g}{\partial \boldsymbol{n}_t})^T \, \frac{\partial \boldsymbol{n}_t}{\partial \boldsymbol{m}_t} + (\frac{\partial g}{\partial \boldsymbol{m}_t})^T)^{(d)} \frac{d m_t^{(d)}}{d \boldsymbol{U}}  \\
&=& 0 + \sum_{t=1}^T \sum_d \boldsymbol{r}_t^{(d)} \, \Big( \boldsymbol{P}_t^{(d)} \Big) \\
\boldsymbol{P}_t^{(d)} &\triangleq& \frac{d \boldsymbol{m}_t^{(d)}}{d \boldsymbol{U}}\\
&=& \frac{\partial \boldsymbol{m}_t^{(d)}}{\partial \boldsymbol{U}} + \sum_{d^\prime} \frac{\partial \boldsymbol{m}_t^{(d)}}{\partial \boldsymbol{m}_{t-1}^{(d^\prime)}} \boldsymbol{P}_{t-1}^{(d^\prime)}\\
&=& \mbox{"Zero matrix except d'th row set to $\boldsymbol{x}_t^T$"} + \sum_{d^\prime} v_{d,d^\prime}  {\widetilde{h}}^\prime (m_{t-1}^{(d^\prime)}) \boldsymbol{P}_{t-1}^{(d^\prime)} \\
\boldsymbol{P}_1^{(d)} &\triangleq& \mbox{"Zero matrix except d'th row set to $\boldsymbol{x}^T_1$"} \,.
\end{eqnarray}

\linesep

\begin{eqnarray}
(\frac{d g}{d \boldsymbol{m}_0})^T &=& (\frac{\partial g}{\partial \boldsymbol{m}_0})^T + \sum_{t=1}^T (\frac{d g}{d \boldsymbol{m}_t})^T \, \frac{d \boldsymbol{m}_t}{d \boldsymbol{m}_0} \, \\
&=& (\frac{\partial g}{\partial \boldsymbol{m}_0})^T + \sum_{t=1}^T ((\frac{\partial g}{\partial \boldsymbol{n}_t})^T \, \frac{\partial \boldsymbol{n}_t}{\partial \boldsymbol{m}_t} + (\frac{\partial g}{\partial \boldsymbol{m}_t})^T) \, \frac{d \boldsymbol{m}_t}{d \boldsymbol{m}_0} \, \\
&=& \lambda \Big( \big( {\widetilde{h^2}}^\prime(\boldsymbol{m}_0) -2 {\widetilde{h}}^\prime(\boldsymbol{m}_0) \odot {\widetilde{h}}(\boldsymbol{m}_0) \big)^T \odot \big( \boldsymbol{1}^T (\boldsymbol{V} \odot \boldsymbol{V}) \big) \\
& & \quad\quad\quad  + 2 H \sigma^2 ({\widetilde{h}}^\prime(\boldsymbol{m}_0) \odot {\widetilde{h}}(\boldsymbol{m}_0)  )^T \Big) \\
& & \quad + \sum_{t=1}^T \boldsymbol{r}_t \, \Big( \boldsymbol{Q}_t \Big) \\
\boldsymbol{Q}_t &\triangleq& \frac{d \boldsymbol{m}_t}{d \boldsymbol{m}_0} = \frac{\partial \boldsymbol{m}_t}{\partial \boldsymbol{m}_{t-1}} \boldsymbol{Q}_{t-1} = \boldsymbol{V} \diag \big( {\widetilde{h}}^\prime (\boldsymbol{m}_{t-1}) \big) \boldsymbol{Q}_{t-1}\\
\boldsymbol{Q}_0 &\triangleq& \boldsymbol{I} \,.
\end{eqnarray}

\linesep

\section{Bounding Linearization Error}
\label{sec:linear}

\paragraph{\bf Proposition 1}
Assume $n \geq 5$, $c_f \geq \frac{1}{2\pi}\sum_{j,k} \| \frac{d^2 f}{d x_j\,  d x_k}\|_{\frac{n}{2}}$ and $\rho^2 c_f \frac{1}{\sigma^2} \leq \epsilon$. Then if follows that $\forall \boldsymbol{x} \,;\, \| \boldsymbol{x} - \boldsymbol{x}_0 \| \leq \rho \Rightarrow | g(\boldsymbol{x}_0) + (\boldsymbol{x} - \boldsymbol{x}_0)^T \nabla g(\boldsymbol{x}_0) - g(\boldsymbol{x})| \leq \epsilon$.

\begin{proof}

First we claim that $\frac{1}{2} \Lambda_g \leq  \frac{1}{2 \pi \sigma^2} \sum_{j,k} \| \frac{d^2 f}{d x_j\,  d x_k}\|_{\frac{n}{2}}$. We prove this claim as below,

\begin{eqnarray}
\frac{1}{2}  \Lambda_g &\leq& \max_{\boldsymbol{x}} \| \nabla^2 g(\boldsymbol{x})\|_F \\
&\leq& \max_{\boldsymbol{x}}  \sum_{j,k} | \frac{d^2 g}{d x_j\,  d x_k}(\boldsymbol{x})| \\
&\leq&   \sum_{j,k} \max_{\boldsymbol{x}} | \frac{d^2 g}{d x_j\,  d x_k}(\boldsymbol{x})|\\
&=& \sum_{j,k} \| \frac{d^2 g}{d x_j\,  d x_k}\|_\infty \\
&=& \sum_{j,k} \| \frac{d^2 f}{d x_j\,  d x_k} \star k_\sigma \|_\infty \\
\label{eq:young}
&\leq& \sum_{j,k} \| \frac{d^2 f}{d x_j\,  d x_k}\|_{\frac{p}{p-1}} \,\| k_\sigma \|_p \\
&\leq& \Big( \sum_{j,k} \| \frac{d^2 f}{d x_j\,  d x_k}\|_{\frac{p}{p-1}} \Big) \Big(\int_\mathcal{X} k^p_\sigma(\boldsymbol{x}) \, d \boldsymbol{x} \Big)^{\frac{1}{p}} \\
&\leq& \Big( \sum_{j,k} \| \frac{d^2 f}{d x_j\,  d x_k}\|_{\frac{p}{p-1}} \Big) \Big(\frac{(2 \pi)^{(1 - p)} \sigma^{2(1 - p)}}{p} \Big)^{\frac{n}{4p}} \\
&=& \Big( \sum_{j,k} \| \frac{d^2 f}{d x_j\,  d x_k}\|_{\frac{p}{p-1}} \Big) \Big(\frac{(2 \pi)^{(1 - p)} }{p} \Big)^{\frac{n}{4p}} \sigma^{\frac{n(1 - p)}{2p}}\,,
\end{eqnarray}

where (\ref{eq:young}) is due to Young's convolution inequality and holds for any $p \geq 1$. In particular, when $n \geq 5$, by setting $p=\frac{n}{n-4}$, we obtain

\begin{eqnarray}
\frac{1}{2}  \Lambda_g &\leq& \Big( \sum_{j,k} \| \frac{d^2 f}{d x_j\,  d x_k}\|_{\frac{p}{p-1}} \Big) \Big(\frac{(2 \pi)^{(1 - p)} }{p} \Big)^{\frac{n}{4p}} \sigma^{\frac{n(1 - p)}{2p}} \\
\frac{1}{2}  \Lambda_g &=& \Big( \sum_{j,k} \| \frac{d^2 f}{d x_j\,  d x_k}\|_{\frac{n}{2}} \Big) \frac{1}{2 \pi \sigma^2} \Big(1-\frac{4}{n} \Big)^{\frac{n}{4}-1} \\
\frac{1}{2}  \Lambda_g &\leq& \Big( \sum_{j,k} \| \frac{d^2 f}{d x_j\,  d x_k}\|_{\frac{n}{2}} \Big) \frac{1}{2 \pi \sigma^2}  \,.
\end{eqnarray}

This proves our earlier claim that $\frac{1}{2} \Lambda_g \leq  \frac{1}{2 \pi \sigma^2} \sum_{j,k} \| \frac{d^2 f}{d x_j\,  d x_k}\|_{\frac{n}{2}}$. Combining this with the assumption $\frac{1}{2\pi} \sum_{j,k} \| \frac{d^2 f}{d x_j\,  d x_k}\|_{\frac{n}{2}} \leq c_f $, it follows that $\frac{1}{2} \Lambda_g \leq c_f \frac{1}{\sigma^2}$, which implies $\frac{1}{2} \rho^2 \Lambda_g \leq \rho^2 c_f \frac{1}{\sigma^2}$. The latter combined with the assumption $\rho^2 c_f \frac{1}{\sigma^2} \leq \epsilon$ yields  $\frac{1}{2} \rho^2 \Lambda_g \leq \epsilon$. Combining this with the Taylor's remainder theorem $| g(\boldsymbol{x}_0) + (\boldsymbol{x} - \boldsymbol{x}_0)^T \nabla g(\boldsymbol{x}_0) - g(\boldsymbol{x})| \leq \frac{1}{2} \rho^2 \Lambda_g$ gives $| g(\boldsymbol{x}_0) + (\boldsymbol{x} - \boldsymbol{x}_0)^T \nabla g(\boldsymbol{x}_0) - g(\boldsymbol{x})| \leq \epsilon$.

\qed
\end{proof}

\end{document}